\documentclass[conference]{IEEEtran}
\IEEEoverridecommandlockouts
\usepackage{cite}
\usepackage{amsmath,amssymb,amsfonts}
\usepackage{algorithmic}
\usepackage{graphicx}
\usepackage{textcomp}
\usepackage{xcolor}

\usepackage{booktabs}
\usepackage{kotex}
\usepackage[ruled,vlined]{algorithm2e}
\usepackage{multirow}
\usepackage{makecell}
\usepackage{url}

\usepackage[table]{xcolor}
\usepackage{booktabs}

\definecolor{rankhue}{HTML}{4A90D9}
\newcommand{\rka}[1]{\cellcolor{rankhue!35}#1}
\newcommand{\rkb}[1]{\cellcolor{rankhue!20}#1}
\newcommand{\rkc}[1]{\cellcolor{rankhue!10}#1}

\newcommand{\oursol}{\textsc{IDTD}}
\newcommand{\spiactive}{\textsc{SPI-Active}}
\newcommand{\tr}{\operatorname{tr}}

\newcommand{\ci}[1]{{\scriptsize$\pm#1$}}

\def\BibTeX{{\rm B\kern-.05em{\sc i\kern-.025em b}\kern-.08em
    T\kern-.1667em\lower.7ex\hbox{E}\kern-.125emX}}
\begin{document}

\title{
Informationally Decoupled Trajectory Design for Sim-to-Real System Identification
}

\author{
\IEEEauthorblockN{Sangwoo Shin\textsuperscript{*}, Ashvin Anilkumar, Ryan Gao, Josiah P. Hanna}
\IEEEauthorblockA{University of Wisconsin--Madison\\
\texttt{sshin237@cs.wisc.edu}\\
\textsuperscript{*}Corresponding author.}
}

\maketitle

\begin{abstract}
Sampling-based system identification estimates physically meaningful parameters by tuning a simulator to reproduce the target system dynamics, providing an interpretable approach to improving sim-to-real transfer.
Yet when the collected trajectories do not distinguish the effects of different parameters, multiple parameter combinations can reproduce those trajectories, leading to unreliable parameter estimates.
%
To address this challenge, we introduce an Informationally Decoupled Trajectory Design framework (\oursol{}),
which formulates the objective for the exploration policy built on the Schur complement score derived from the Fisher information matrix.
To faithfully reflect parameter separability in the exploration objective, \oursol{} normalizes the score against the per-parameter information, 
selects the most favorable trajectory segment for each parameter, and aggregates the resulting scores logarithmically.
The optimized trajectory is therefore composed of complementary intervals, each exposing a distinct subset of parameters whose contribution to the motion over that interval can be attributed unambiguously.
%
Across diverse simulation environments, ranging from a linear-dynamics system to the Go2 quadruped, G1 humanoid, and Crazyflie quadrotor, \oursol{} reduces the parameter identification error by $39.6\%$ on average relative to the strongest prior active-exploration baseline and attains improved downstream policy transfer. 
Furthermore, we validate the proposed trajectory design on a real K1 humanoid, demonstrating that the resulting identified parameters accurately capture the real-system dynamics.
Code is available at \url{https://github.com/jsw7460/IDTD}.

\end{abstract}



\section{Introduction}
Reinforcement learning (RL) in massively parallel simulation has become 
a widely adopted approach
for learning-based robot control, yet policies trained in simulation degrade on hardware whenever the simulated dynamics deviate from the target physical system~\cite{hwangbo2019learning, sim2real_survey}.
A common remedy is domain randomization, which trains a policy over a distribution of physical parameters, such as inertial properties, actuator gains, and friction, so that the real system falls within its support~\cite{tobin2017domain, peng2018sim}.
This approach, however, requires covering a wide range of parameters, which tends to produce conservative policies that sacrifice performance for robustness.

System identification (SysID) offers an alternative approach.
Rather than training against a broad range of possible dynamics, SysID explicitly estimates the physical parameters that make the simulator reproduce the target system dynamics.
Within SysID, sampling-based methods have recently gained traction~\cite{spi_active, asid}, as massively parallel GPU simulation makes black-box search over physical parameters practical.
These methods evaluate candidate parameter vectors through forward dynamics simulation and search for the parameters under which the simulated trajectories best match the recorded real-world trajectories.
By avoiding differentiation through the simulator, sampling-based SysID can directly leverage modern physics simulators with contact and actuator dynamics~\cite{mujoco, isaac, newton, genesis} while retaining physically interpretable parameters.

While sampling-based approaches make the parameter search practical, their success depends critically on solving the complementary problem of determining \textit{which trajectories should be collected from the target system for reliable identification.}
Prior work addresses this question by seeking informative trajectories, where informativeness is quantified through the Fisher Information Matrix (FIM), whose inverse lower-bounds the covariance of the parameter estimate~\cite{cramer-rho}.
Based on this bound, active exploration methods search for a policy that collects trajectories minimizing the trace of the inverse FIM, aggregating the estimation uncertainty across all parameters into a single objective~\cite{asid, spi_active}.
However, this criterion does not treat the amount of information and the separability of the parameters as distinct quantities, so increased excitation can improve the aggregate criterion even when some parameter effects become less distinguishable.
%
%
Consequently, errors in different parameters can compensate for one another, allowing simulations using the identified parameters to closely match the real-world trajectories even when the individual parameters differ substantially from their true values.
%
For instance, consider identifying the mass $m$ and the damping $c$ of a one-dimensional system, where a force is applied to a point mass.
A trace-based excitation makes the mass move faster and more abruptly, since larger, more energetic motion carries more total information about the parameters.
Under such motion, however, the sensitivities to $m$ and $c$ stay nearly collinear, so increasing the mass while reducing the damping produces nearly the same trajectory under the same control input.
The criterion therefore keeps improving even though the two parameters remain difficult to tell apart.
In this simple example, the excitation can be manually designed to alternate between acceleration-dominant ($c\dot{x}\approx0$) and velocity-dominant ($m\ddot{x}\approx0$) phases to distinguish the mass from the damping.
Extending such parameter-specific excitation to the dozens of coupled parameters in modern robot models, however, is highly challenging, as each parameter may require a different excitation regime.

To address this challenge, we present \textbf{I}nformationally \textbf{D}ecoupled \textbf{T}rajectory \textbf{D}esign (\oursol), which searches for motions under which every parameter leaves a unique effect on the trajectory that no other parameter can reproduce.
Specifically, 
\oursol{} builds on a per-parameter score derived from the Schur complement of the Fisher information, which measures how much of a parameter's effect on the motion remains after the effects of all other parameters are accounted for.
A parameter whose effect can be reproduced by the others scores near zero, while a parameter with a distinct effect scores high; a high score thus means the parameter is cleanly distinguishable, or decoupled, from the rest.
%
%
Furthermore, to prevent the objective from being improved through excitation alone, for instance by moving a few responsive joints more aggressively without better separating their parameter effects, \oursol{} normalizes each score by the information carried by the corresponding parameter. 
It then evaluates the normalized scores across multiple trajectory segments to encourage complementary excitation of different parameters throughout the trajectory.
%
%
As a result, \oursol{} designs trajectories that expose every parameter somewhere along the motion rather than accumulating information for only a small subset.

We evaluate \oursol{} in simulation on a linear-dynamics system, Unitree Go2, Unitree G1, and Crazyflie, demonstrating improved parameter identification, target-system prediction, and downstream policy transfer.
We further validate \oursol{} on a real Booster K1 humanoid, where the identified parameters more accurately capture the target-system dynamics.

The main contributions of our work are as follows.
\textbf{(i)} We address the challenging yet practical problem of designing exploration trajectories that distinguish the effects of physical parameters, enabling reliable sampling-based system identification.
\textbf{(ii)} We propose \oursol{}, a trajectory design framework that explicitly balances information quantity with parameter distinguishability through a normalized Schur-complement score and parameter-wise trajectory coverage.
\textbf{(iii)} We demonstrate that \oursol{} improves parameter identification accuracy and downstream policy transfer across simulated Unitree Go2, Unitree G1, and Crazyflie platforms, and better captures real-world dynamics on a Booster K1 humanoid.

\section{Related Works}

\subsection{Sim-to-Real Transfer and System Identification}
Reducing the sim-to-real gap has been approached from two complementary directions.
Domain randomization exposes the policy to a wide distribution of dynamics so that the real system falls within the training support~\cite{tobin2017domain, peng2018sim}; the resulting robustness, however, comes at the cost of conservative behavior, and the randomization ranges themselves require tuning.

System identification instead estimates the physical parameters of the specific robot and its environment from real-world data.
Classical approaches exploit the linearity of rigid-body dynamics in the inertial parameters and estimate them by least squares over hand-designed excitation~\cite{atkeson1986estimation, gautier1992exciting}, but they require an explicit structured dynamics model and do not scale to the contact-rich, hard-to-model effects of real robots.

As massively parallel simulation has become fast and reliable, system identification is increasingly performed by tuning simulator parameters until simulated trajectories match real trajectories. 
One line of work makes the forward dynamics differentiable with respect to the parameters, so the identification objective can be minimized by backpropagating gradients directly through the simulation~\cite{chen2022realtime, contactexcitation}. Its reliance on smoothed contact models, however, limits its scalability to contact-rich tasks. 
Recent work therefore favors sampling-based identification, which pairs the simulator with a black-box search algorithm and sidesteps differentiation while keeping the estimated parameters physically interpretable~\cite{spi_active, asid}.

Our work follows the sampling-based paradigm but addresses the complementary question of how exploration should be designed so that every parameter leaves a distinct effect on the resulting trajectory, enabling reliable system identification.

\subsection{Active Exploration and Optimal Experimental Design}
The choice of excitation has long been recognized as central to identification quality. 
In the classical context, exciting trajectories are optimized offline against criteria defined on the regressor matrix, which linearly relates the unknown dynamic parameters to the measured forces for a given trajectory.
A common objective is to minimize the condition number of the regressor, producing excitations that yield parameter estimates more robust to measurement noise~\cite{swevers1997optimal, gautier1992exciting}.
The objective is inherently coupling-aware, since an ill-conditioned regressor indicates that different parameters influence the measurements along nearly identical directions.

For policy-driven system identification, where the excitation is produced by a closed-loop policy, \spiactive{} and ASID~\cite{spi_active, asid} optimize the A-optimal Fisher-information criterion $\tr(F^{-1})$ to encourage informative trajectories.
Contact-implicit trajectory optimization likewise maximizes $\tr({F})$ over excited contact modes~\cite{contactexcitation}.
Optimal experimental design offers related criteria, including D-optimality ($\log\det (F)$) for reducing the overall uncertainty of the parameter estimate and E-optimality ($\lambda_{\min}(F)$, the smallest eigenvalue of $F$) for improving the least informative direction~\cite{pukelsheim2006optimal, dozzi2026}.
In the transition from regressor-based trajectory design to sampling-based identification, the Fisher information has been carried over as the measure of a trajectory, 
whereas parameter separation is addressed only implicitly through aggregate objectives.
The Schur complement provides an explicit characterization of parameter separation by quantifying the information that remains for a parameter subset after accounting for correlations with the remaining parameters~\cite{ljung1999system}.
QOED~\cite{qoed} leverages this quantity to improve the distinguishability of a subset of task-relevant parameters, leading to improved sim-to-real policy transfer.
However, system identification requires all physical parameters to be identifiable, rather than only a selected subset.

To the best of our knowledge, \oursol{} is the first exploration objective for sampling-based system identification that uses per-parameter separability to enforce coverage of all physical parameters.
%

\section{Preliminaries}
\subsection{Sampling-Based System Identification}
\label{subsec: prelim_sysid}
Let $\theta \in \mathbb{R}^{N}$ denote the vector of parameters to identify (e.g., base inertial properties, per-joint PD gains, and friction coefficients), and let $\pi$ be a policy pretrained in simulation, conditioned on a command $c_t$ (e.g., base velocity and gait parameters).
A dataset $\mathcal{D} = \{\tau = (s_0, a_0, s_1, \cdots) \}$ of trajectories is collected on the target system by rolling out $\pi$ under a command sequence $\mathbf{c} = (c_1, \ldots, c_T)$, recording states and executed actions.
In this paper, we focus on a floating-base robot policy $\pi$, where the state $s_t$ comprises measurable proprioceptive signals, including IMU readings, joint positions and velocities from the encoders, and, when available, joint torques.
Sampling-based system identification then solves
\vspace{-4pt}
\begin{equation}
\label{eqn: sysid_objective}
    \hat{\theta} = \mathop{\textnormal{argmin}}_{\theta} \; \sum_{\tau \in \mathcal{D}} \sum_{t=1}^T \big\| f(s_t, a_t; \theta) - s_{t+1} \big\|^2,
\vspace{-4pt}
\end{equation}
where $f(\cdot \,; \theta)$ denotes a forward simulation step under parameters $\theta$, and $a_t$ is the action recorded in trajectory $\tau$.
For parameter estimation, we treat the simulator as a black box and optimize Eq.~\eqref{eqn: sysid_objective} with CMA-ES~\cite{cmaes}.

\begin{figure}[t]
\centering
\includegraphics[width=\columnwidth]{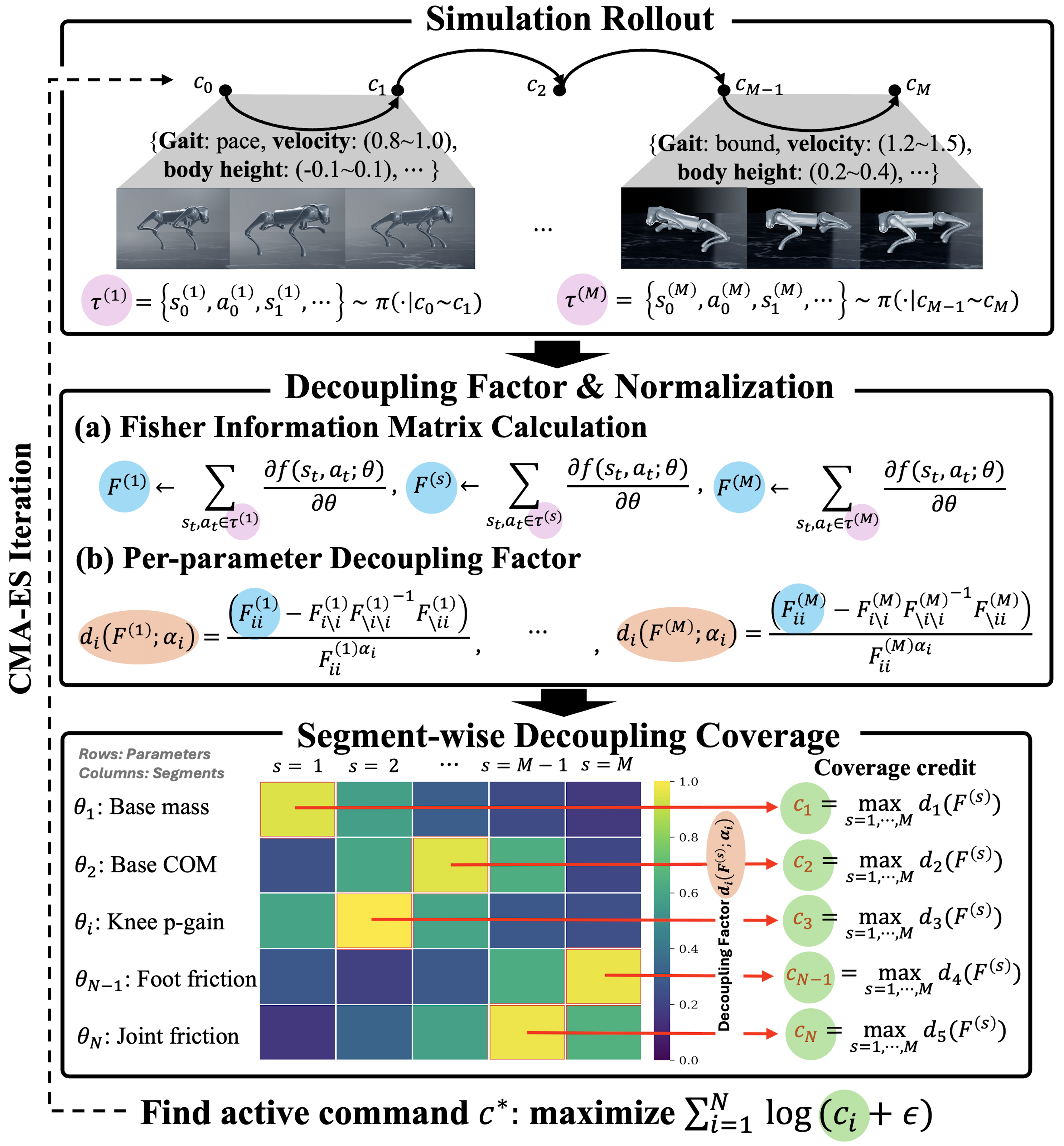}
\vspace{-20pt}
\caption{Overview of \oursol{}. (1)~A command sequence is rolled out in simulation over $M$ segments. (2)~Each segment yields a Fisher information $F^{(s)}$, from which the per-parameter decoupling factor $d_i(F^{(s)};\alpha_i)$ follows through the Schur complement and the normalization of Sec.~\ref{subsec: decoupling_factor}. (3)~Each parameter takes the score $c_i$ of its best segment, and \oursol{} optimizes the logarithmic sum of these scores.}
\vspace{-16pt}
\label{fig:overview}
\end{figure}

\subsection{Fisher Information for Trajectory Design}
\label{subsec: prelim_fim}
The informativeness of a trajectory is measured by the Fisher Information Matrix (FIM). By the Cram\'er--Rao bound~\cite{cramer-rho}, the covariance of unbiased estimator $\hat{\theta}$ of the true parameters $\theta^\star$ is lower-bounded by the inverse of the FIM,
\vspace{-2pt}
\begin{equation}
\label{eqn: cramer_rao}
    \mathbb{E}_{s_{1:T} \sim p(\cdot \mid \theta^\star)}\big[(\hat{\theta} - \theta^\star)(\hat{\theta} - \theta^\star)^\top\big] \succeq F(\theta^\star)^{-1}.
\vspace{-2pt}
\end{equation}
This bound provides a classical interpretation of the FIM as a measure of parameter identifiability and motivates its use as a trajectory-design criterion.

%
Assuming the stochastic dynamics 
$s_{t+1} = f(s_t, a_t ; \theta) + w_t$ with additive Gaussian process noise $w_t \sim \mathcal{N}(0, \sigma^2I)$,
the FIM of a trajectory is approximated as~\cite{spi_active}:
\vspace{-1pt}
\begin{equation}
\label{eqn: fim}
    F(\theta, \mathbf{c}) = \sigma^{-2} \sum_{t=1}^{T} J_t^\top J_t, \qquad J_t = \frac{\partial f(s_t, a_t; \theta)}{\partial \theta},
\vspace{-1pt}
\end{equation}
where $J_t$ is the sensitivity of the simulation dynamics to the parameters along the trajectory induced by the policy under the exploration command sequence $\mathbf{c}$.
Because evaluating FIM in Eq.~\eqref{eqn: cramer_rao} requires the true parameter $\theta^\star$, which is unavailable in practice, 
we instead evaluate the FIM at a nominal parameter value $\theta_0$.
To avoid altering the behavior of the pretrained policy for safe data collection on the target system, we optimize the command sequence while keeping the policy fixed.
Active exploration~\cite{asid, spi_active} searches for a command that minimizes the aggregate uncertainty,
\begin{equation}
\label{eqn: spi_objective}
    \mathbf{c}^{\star} = \mathop{\textnormal{argmin}}_{\mathbf{c}} \; \tr \big( F(\theta_0, \mathbf{c})^{-1} \big).
\end{equation}
The command sequence $\mathbf{c}^{\star}$ optimized in simulation is then executed on the target system to collect a trajectory, from which the parameters are estimated through Eq.~\eqref{eqn: sysid_objective}.

\subsection{Problem Statement}
Given a command-conditioned pretrained policy $\pi$, a nominal parameter estimate $\theta_0$, and a fixed data-collection time budget, our goal is to design an exploration objective, in place of the criterion of Eq.~\eqref{eqn: spi_objective}, that enables accurate recovery of all physical parameters.

\section{Methods}
\subsection{Motivation: Parameter Coupling in System Identification}
\label{subsec: motivation}
We analyze a two-parameter linear system to illustrate how parameter
coupling allows estimation errors in one parameter to be absorbed by
another, motivating an explicit characterization of parameter separability.
Consider a scalar state governed by a discrete-time mass-damper system, in which the control input $u_t$ drives the state through a compliance $\theta^\star$ and the velocity $\dot{x}_t$ resists it through a damping $\phi^\star$,
\begin{equation}
\vspace{-1pt}
\label{eqn: linear_dynamics}
    x_{t+1} = x_t + \theta^\star u_t - \phi^\star \dot{x}_t + w_t,
    \qquad w_t \sim \mathcal{N}(0, \sigma^2),
\vspace{-1pt}
\end{equation}
where $w_t$ is i.i.d. Gaussian noise.
By Eq.~\eqref{eqn: fim}, the Fisher information matrix $\mathcal{I}$ for the two parameters has entries
\begin{equation}
\label{eqn: linear_fim}
    \mathcal{I}_{\theta\theta} = \frac{1}{\sigma^2}\sum_{t} u_t^2,
    \
    \mathcal{I}_{\phi\phi} = \frac{1}{\sigma^2}\sum_{t} \dot{x}_t^2,
    \
    \mathcal{I}_{\theta\phi} 
    = -\frac{1}{\sigma^2}\sum_{t} u_t \dot{x}_t,
\end{equation}
where $\mathcal{I}_{\theta\theta}$ and $\mathcal{I}_{\phi\phi}$ measure the information about each parameter individually, while the off-diagonal $\mathcal{I}_{\theta\phi}$ measures their coupling along the trajectory.
To understand how an error in one parameter propagates to another, suppose that the damping parameter has been estimated as $\hat{\phi}$. 
Instantiating Eq.~\eqref{eqn: sysid_objective} and conditioning on $\hat{\phi}$, the least-squares estimate of $\theta$ has the closed form
\vspace{-2pt}
\begin{equation}
\label{eqn: ls_theta}
    \hat{\theta} = \frac{\sum_t u_t(x_{t+1}-x_t+\hat{\phi}\dot{x}_t)}{\sum_t u_t^2}.
\vspace{-2pt}
\end{equation}
Substituting the true dynamics for $x_{t+1}$ in Eq.~\eqref{eqn: linear_dynamics} gives
\begin{equation}
\label{eqn: ls_expanded}
    \hat{\theta} = \theta^\star + (\hat{\phi}-\phi^\star )\,\frac{\sum_t u_t \dot{x}_t}{\sum_t u_t^2} + \frac{\sum_t u_t w_t}{\sum_t u_t^2}.
\end{equation}
Taking the expectation over the noise eliminates the last term, and the remaining ratio equals $-\mathcal{I}_{\theta\phi}/\mathcal{I}_{\theta\theta}$ by Eq.~\eqref{eqn: linear_fim}, so
\vspace{-2pt}
\begin{equation}
\label{eqn: absorption_bias}
    \mathbb{E}[\hat{\theta}] - \theta^\star
    = -\,(\hat{\phi} - \phi^\star) \cdot \frac{\mathcal{I}_{\theta\phi}}{\mathcal{I}_{\theta\theta}}.
\vspace{-2pt}
\end{equation}
For $\hat{\theta}$ to recover $\theta^\star$, Eq.~\eqref{eqn: absorption_bias} requires either (i) the nuisance estimate to be exact, $\hat{\phi} = \phi^\star$, or (ii) the coupling ratio $\mathcal{I}_{\theta\phi}/\mathcal{I}_{\theta\theta}$ to vanish. Since $\phi$ is itself estimated and cannot be assumed to match $\phi^\star$, the only quantity under control is the coupling ratio. 
Reducing this ratio requires suppressing the off-diagonal term $\mathcal{I}_{\theta\phi}$, not merely enlarging the diagonal information $\mathcal{I}_{\theta\theta}$. 
The smaller the off-diagonal coupling, the less the error in $\hat{\phi}$ leaks into $\hat{\theta}$, and the more accurately $\theta$ is identified.
This coupling can be characterized through the Schur complement of the Fisher information, which measures the information about $\theta$ that remains after accounting for the effect of $\phi$,
\vspace{-2pt}
\begin{equation}
\label{eqn: schur_motivation}
    S_\theta = \mathcal{I}_{\theta\theta} - \mathcal{I}_{\theta\phi}\,\mathcal{I}_{\phi\phi}^{-1}\,\mathcal{I}_{\phi\theta}.
\vspace{-2pt}
\end{equation}
%
%
The Schur complement $S_\theta$ reflects both the information about $\theta$ and its separability from $\phi$, with the latter directly related to the bias that causes the error absorption in Eq.~\eqref{eqn: absorption_bias}.

\subsection{Per-Parameter Decoupling Factor}
\label{subsec: decoupling_factor}

We now extend the two-parameter analysis to the full parameter vector $\theta = (\theta_1, \cdots, \theta_N) \in \mathbb{R}^N$. For each parameter $\theta_i$, we partition $\theta = (\theta_i, \theta_{\setminus i})$ into the target $\theta_i$ 
and the remaining parameters $\theta_{\setminus i}$, 
and partition the Fisher information $F$ accordingly into the blocks 
$F_{ii}\in\mathbb{R}^{1\times1}$,
$F_{i,\setminus i}\in\mathbb{R}^{1\times(N-1)}$,
and
$F_{\setminus i,\setminus i}\in\mathbb{R}^{(N-1)\times(N-1)}$.
The information about $\theta_i$ that remains after accounting for the remaining parameters is given by the Schur complement
\vspace{-2pt}
\begin{equation}
\label{eqn: schur_single}
    S_i(F) = F_{ii} - F_{i,\setminus i}\, F_{\setminus i,\setminus i}^{-1}\, F_{\setminus i, i} \;\in\; \mathbb{R}.
\vspace{-2pt}
\end{equation}
A parameter whose sensitivity is orthogonal to the remaining parameters attains $S_i = F_{ii}$, while a parameter fully absorbed by the others has $S_i \to 0$.

The Schur complement also clarifies the relation to the A-optimal
criterion of Eq.~\eqref{eqn: spi_objective}, since
$\operatorname{tr}(F^{-1}) = \sum_{i=1}^{N} 1/S_i(F)$.
Although A-optimal design is not formulated in terms of parameter
separability, its aggregate uncertainty criterion implicitly reflects
parameter coupling.
In contrast, \oursol{} explicitly uses the Schur complement to quantify
the separability of each parameter, and introduces mechanisms to
control its dependence on information quantity and ensure coverage
across parameters.
We next describe how these mechanisms are incorporated into our exploration objective.


\subsection{Two Normalizations for Faithful Decoupling}
\label{subsec: normalization}

\noindent\textbf{Sensitivity normalization.}
Computing the Schur complement requires the Fisher information $F$, which by Eq.~\eqref{eqn: fim} is built from the sensitivity of the forward dynamics to the parameters, $J_t = \partial f(s_t, a_t; \theta) / \partial \theta$.
In simulation this derivative is unavailable in closed form and is estimated by finite differences, perturbing each parameter by a step $\delta\theta_i$ and measuring the resulting change in the predicted next state.
The parameters, however, span several orders of magnitude in natural scale, so a single shared step distorts the resulting sensitivities.
With an absolute step $\delta\theta_{\text{fric}} = \delta\theta_{k_p} = 10^{-3}$, for example, the friction coefficient is perturbed by roughly $0.2\%$ of its value while the proportional gain is perturbed by only $0.001\%$, so the finite-difference FIM overstates the information about friction and understates it for the gain purely as an artifact of the step size.
We therefore scale the step of each parameter to its search range, $\delta\theta_i \propto \text{range}(\theta_i)$, so that every parameter is perturbed by a comparable relative amount and every column of $J_t$ is evaluated consistently.

\noindent\textbf{Decoupling normalization.}
The raw Schur complement combines two quantities that should be treated separately. Specifically, factoring Eq.~\eqref{eqn: schur_single},
\vspace{-4pt}
\begin{equation}
\label{eqn: schur_factored}
    S_i(F)
    =
    \underbrace{F_{ii}}_{\text{information quantity}}
    \underbrace{\left(
    1 -
    \frac{
        F_{i,\setminus i}\,
        F_{\setminus i,\setminus i}^{-1}\,
        F_{\setminus i,i}
    }{F_{ii}}
    \right)}_{\text{decoupling ratio}},
\vspace{-4pt}
\end{equation}
splits it into the diagonal information $F_{ii}$, which measures excitation
strength, and the parenthesized decoupling ratio in $[0,1]$, which measures
the fraction of information about $\theta_i$ that remains after accounting
for the other parameters and corresponds to the classical inverse variance
inflation factor $1/\mathrm{VIF}_i$~\cite{marquardt1970}.
However, the magnitude of $F_{ii}$ varies by orders of magnitude across parameters, so a large $F_{ii}$ can yield a large $S_i$ even when the decoupling ratio remains small.
Consequently, maximizing $\sum_i S_i$ can favor strongly excited parameters even when their effects remain poorly separated.
We therefore damp the diagonal factor using a parameter-specific exponent $\alpha_i$ and define the decoupling factor
\vspace{-4pt}
\begin{equation}
\label{eqn: decoupling_factor}
d_i(F; \alpha_i)
=
\frac{S_i(F)}{F_{ii}^{\alpha_i}}
=
F_{ii}^{1-\alpha_i}
\left(
1 -
\frac{
F_{i,\setminus i}
F_{\setminus i,\setminus i}^{-1}
F_{\setminus i,i}
}{F_{ii}}
\right).
\end{equation}
The exponent $\alpha_i$ controls how much of the diagonal magnitude is removed. At $\alpha_i = 1$, the diagonal factor cancels and $d_i \in [0,1]$ becomes the pure decoupling ratio, independent of excitation strength.
At $\alpha_i = 0$, no damping is applied and $d_i = S_i$ reverts to the diagonal-dominated regime above.
In our experiments, we use parameter-specific $\alpha_i \geq 0.7$.

{
\setlength{\textfloatsep}{0pt}
\begin{algorithm}[t]
\caption{Trajectory design in \oursol{}}
\label{alg:idtd}
\small
\begin{algorithmic}[1]
\REQUIRE policy $\pi$, initial $\theta_0 \in \mathbb{R}^N$, segments $M$, exponents $\alpha_i$
\ENSURE optimized excitation command $\mathbf{c}^\star$
\WHILE{CMA-ES not converged}
    \STATE sample population $\{\mathbf{c}^{(p)}\}_{p=1}^{P}$
    \FOR{each candidate $\mathbf{c}^{(p)}$}
        \STATE trajectory $\tau \sim \pi(\cdot \mid \mathbf{c}^{(p)})$ at $\theta_0$
        \FOR{$i = 1$ to $N$}
            \STATE trajectory  $\tau_i \sim \pi(\cdot \mid \mathbf{c}^{(p)})$ at $\theta_0 + \delta\theta_i \mathbf{e}_i$
            \STATE $J^{(i)} \leftarrow (\tau_i - \tau) / \delta\theta_i$
        \ENDFOR
        \STATE $J_t \longleftarrow [J^{(1)}_t, \dots, J^{(N)}_t]$ for each timestep $t$ in $\tau$
        \STATE $F^{(s)} \longleftarrow \sigma^{-2}\sum_{t \in \text{seg } s} J_t^\top J_t$ for each segment $s$
        \STATE $c_i \leftarrow \max_s \, S_i(F^{(s)}) / [F_{ii}^{(s)}]^{\alpha_i}$ for all $i$
        \STATE $\mathcal{J}^{(p)} \leftarrow \sum_i \log(c_i + \varepsilon)$
    \ENDFOR
    \STATE update CMA-ES with $\{\mathcal{J}^{(p)}\}$ 
\ENDWHILE
\STATE \textbf{return} $\mathbf{c}^\star \leftarrow \arg\max_p \mathcal{J}^{(p)}$
\end{algorithmic}
\end{algorithm}
}
\subsection{Segment-Wise Decoupling Coverage}
\label{subsec: coverage}

Different parameters often require different excitation regimes, making it difficult for a single motion to decouple all parameters simultaneously.
We therefore partition the exploration trajectory into $M$ equal-length segments.
For each segment $s \in \{1, \cdots, M \}$, we compute the corresponding FIM $F^{(s)}$ and the associated decoupling factor.

\noindent\textbf{Coverage score.}
We assign each parameter the highest decoupling score across all segments,
\vspace{-4pt}
\begin{equation}
\label{eqn: coverage_credit}
    c_i = \max_{s \in \{1, \ldots, M\}} d_i(F^{(s)}; \alpha_i).
\vspace{-4pt}
\end{equation}
The $\max$ over segments assigns each parameter to the segment in which it is most cleanly separated, which allows different segments to specialize in different subsets of parameters.

\noindent\textbf{IDTD objective.}
In place of Eq.~\eqref{eqn: spi_objective}, $\oursol$ maximizes the logarithmic sum of the per-parameter coverage scores defined in Eq.~\eqref{eqn: coverage_credit},
\vspace{-8pt}
\begin{equation}
\label{eqn: coverage_objective}
    \mathbf{c}^{\star} = \arg\max_{\mathbf{c}} \; \sum_{i=1}^{N} \log\big(c_i + \varepsilon\big),
\vspace{-4pt}
\end{equation}
with a small $\varepsilon$ for numerical stability.
Under the logarithm, a parameter with $c_i \to 0$, meaning that $\theta_i$ is not decoupled in any segment, drives the objective toward $-\infty$. 
This formulation emphasizes poorly covered parameters while still rewarding improvements across all parameters.
The objective therefore encourages different trajectory segments to provide complementary information, such that every parameter is well identified in at least one segment.
The complete trajectory-design pipeline of $\oursol$ is summarized in Fig.~\ref{fig:overview}, and the corresponding optimization procedure is given in Algorithm~\ref{alg:idtd}.

\section{Experiments}
We evaluate \oursol{} on the linear-dynamics system, in physics-based simulation, and on a real Booster K1 robot.

\subsection{Linear-Dynamics System}
\label{subsec:linear_exp}
To validate the analysis of Sec.~\ref{subsec: motivation}, we manually construct two excitation trajectories on the mass-damper system that differ in how strongly the two parameters are coupled.
Specifically, we collect sequences of states $x_t$ and inputs $u_t$ following the dynamics in Eq.~\eqref{eqn: linear_dynamics}.
The \emph{coupled} trajectory applies a sustained positive input, producing large $\sum_t u_t^2$ and $\sum_t \dot{x}_t^2$ while also inducing strong coupling through the cross term $\sum_t u_t\dot{x}_t$.
The \emph{decoupled} trajectory alternates between high- and low-input phases, maintaining strong excitation while suppressing $\sum_t u_t\dot{x}_t$ by alternating between excitation regimes that predominantly excite $\theta$ and $\phi$, respectively.

\begin{figure}[t]
\centering
\includegraphics[width=\columnwidth]{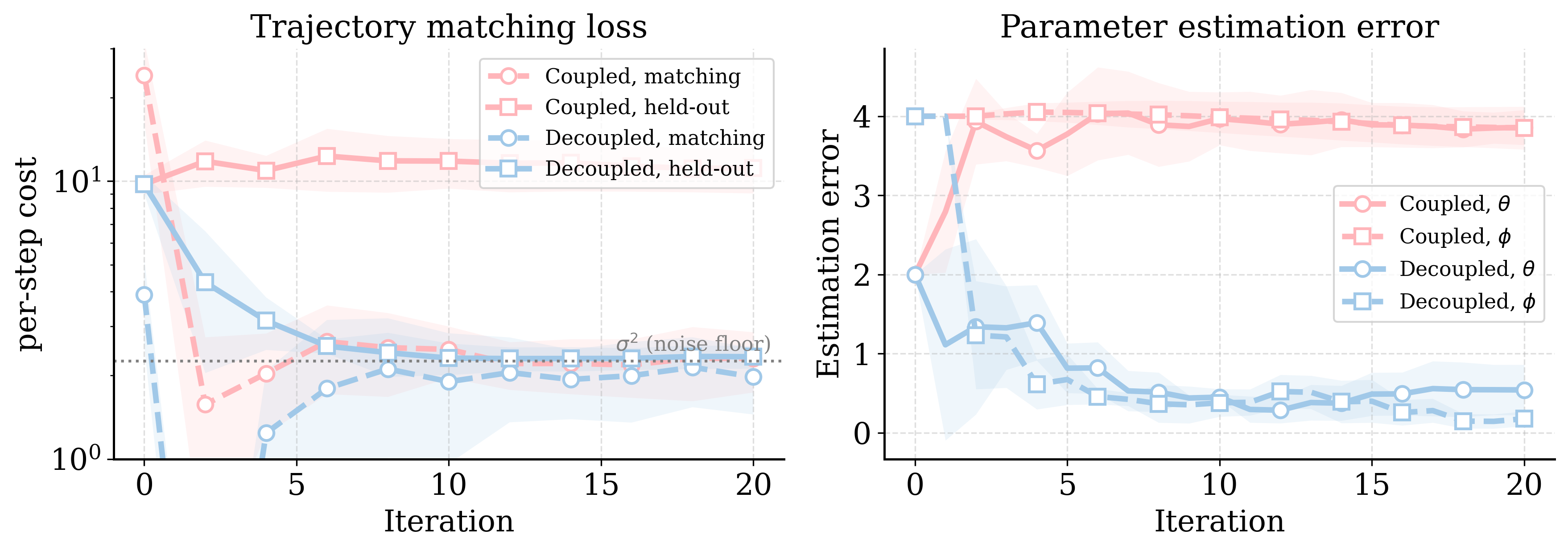}
\vspace{-20pt}
\caption{Trajectory matching loss and parameter estimation error over CMA-ES iterations. 
Shaded regions indicate $95\%$ confidence intervals.}
\label{fig: linear_dynamics}
\vspace{-12pt}
\end{figure}

Figure~\ref{fig: linear_dynamics} plots the trajectory matching costs and the absolute parameter-estimation errors over CMA-ES iterations.
Both trajectories converge to a matching cost close to the noise floor on the identification data, indicating that the observed trajectory can be reproduced almost equally well in both cases.
However, the coupled trajectory retains large parameter-estimation errors and exhibits substantially higher prediction error on the held-out trajectory, whereas the decoupled trajectory accurately recovers both parameters and predicts the held-out trajectory close to the noise floor.
This result shows that a strongly coupled trajectory can be explained by an incorrect parameter combination, making a low trajectory matching cost insufficient evidence of accurate system identification and confirming the importance of suppressing parameter coupling for accurate identification.

\subsection{Physics-Based Simulation}
\label{subsec:sim_setup}
\noindent\textbf{Environments and parameters.}
We evaluate in Newton simulation~\cite{newton}, where the ground-truth parameters are available.
This setting enables direct measurement of identification accuracy and isolates the effect of trajectory design from real-world uncertainties.
We use the Unitree Go2 quadruped, the Unitree G1 humanoid, and the Crazyflie quadrotor.

For the Go2, we identify 50 parameters comprising the base inertial parameters in a log-Cholesky parameterization~\cite{log_cholesky, spi_active, inertia}, per-joint proportional and derivative gains, per-joint Coulomb friction torques, and per-foot contact friction coefficients.
For the G1, we identify 68 parameters comprising the base inertial parameters and per-joint gains.
For the Crazyflie, we identify 17 parameters comprising thrust coefficients, moment coefficients, motor time constants, diagonal inertia terms, and center-of-mass offsets.

\noindent\textbf{Baselines.}
We compare \oursol{} against four trajectory-design methods that differ in the command sequences used for data collection.
\textbf{\spiactive{}}~\cite{spi_active}, which minimizes $\tr(F^{-1})$ as defined in Eq.~\eqref{eqn: spi_objective};
\textbf{\textsc{E-Optimal}}, which maximizes the minimum eigenvalue $\lambda_{\min}(F)$ of the FIM;
\textbf{\textsc{QOED}}~\cite{qoed}, which selects a subset of identifiable parameters through an eigenspace analysis of the FIM, treats the remaining parameters as nuisance variables, and maximizes the trace of the Schur complement of the selected block;
\textbf{\textsc{Passive}}, which reuses the nominal commands, moving in each direction for a fixed duration without active trajectory optimization.
For all methods, the collected trajectories are passed to the same CMA-ES-based system identification procedure, which minimizes Eq.~\eqref{eqn: sysid_objective}.

\noindent\textbf{\oursol{} Implementation.}
Since optimizing the command at every timestep using Eq.~\eqref{eqn: coverage_objective} is intractable, we optimize a small set of control points and recover the command sequence by resampling a degree-$5$ B\'ezier curve fitted through them, following~\cite{spi_active}.
Each data collection episode lasts $10$ seconds and is divided into $M=5$ segments of $2$ seconds, and a separate set of control points is optimized for each segment, so that different segments are free to specialize in different subsets of parameters.
The command space of the pretrained policy bounds the range of motions over which the trajectory design can search.
For the Go2, we therefore adopt the multiplicity-of-behavior formulation of Walk These Ways~\cite{walk_these_ways} and train a policy conditioned on base velocity together with gait frequency, gait phase, foot swing height, and body height, so that varying the command yields qualitatively distinct locomotion behaviors.
For the G1 and the CrazyFlie we keep the standard command spaces, a velocity-tracking policy conditioned on the base linear and angular velocity for the G1 and a position-tracking policy conditioned on a target position for the CrazyFlie.

For trajectory optimization, we use CMA-ES with an initial standard
deviation of $0.25$ and optimize for at most $3000$ generations, with
early stopping when the objective variation falls below $10^{-3}$.
The population sizes are $20$, $12$, and $20$ for Go2, G1, and
Crazyflie, respectively.
For finite-difference sensitivity estimation, we evaluate eight
parallel groups for each candidate and periodically synchronize the
perturbed environments with the nominal environment every five
simulation steps.
We regularize the Fisher submatrices used in the Schur complement by
$10^{-6}I$ and use $\varepsilon=10^{-6}$ in
Eq.~\eqref{eqn: coverage_objective}.
Additional implementation details, including parameter-specific
perturbation steps and decoupling exponents, are provided in the
Appendix.

\begin{table}[t]
\centering
\caption{Normalized parameter-identification error (\%). \\ 
Mean $\pm$ $95\%$ confidence interval, darker = higher rank.}
\label{tab:accuracy_all}
\setlength{\tabcolsep}{2.8pt}
\scriptsize
\begin{tabular}{lccccc}
\toprule
Group & \textbf{IDTD} & \textbf{SPI-Active} & \textbf{E-Opt} & \textbf{QOED} & \textbf{Passive} \\
\midrule
\multicolumn{6}{c}{\textbf{Go2 (50D)}}\\
\midrule
$k_p$    & \rka{$\mathbf{5.23}$\,\ci{0.7}} & \rkb{$6.41$\,\ci{0.4}} & \rkc{$9.07$\,\ci{0.8}} & $10.38$\,\ci{0.6} & $9.19$\,\ci{0.6} \\
$k_d$    & \rkb{$6.71$\,\ci{0.7}} & \rka{$\mathbf{6.55}$\,\ci{0.4}} & $12.00$\,\ci{1.0} & $113.48$\,\ci{36.9} & \rkc{$7.79$\,\ci{0.5}} \\
Joint    & \rka{$\mathbf{10.99}$\,\ci{1.4}} & \rkb{$16.89$\,\ci{1.1}} & $26.43$\,\ci{3.1} & $34.67$\,\ci{2.5} & \rkc{$21.06$\,\ci{0.9}} \\
Foot     & \rkc{$5.90$\,\ci{0.9}} & $8.69$\,\ci{0.9} & $7.75$\,\ci{1.3} & \rkb{$5.18$\,\ci{0.5}} & \rka{$\mathbf{5.09}$\,\ci{0.3}} \\
Base     & \rka{$\mathbf{7.71}$\,\ci{1.3}} & \rkc{$12.85$\,\ci{1.4}} & \rkb{$12.44$\,\ci{2.2}} & $19.51$\,\ci{1.0} & $12.86$\,\ci{1.1} \\
All      & \rka{$\mathbf{7.31}$\,\ci{0.6}} & \rkb{$10.28$\,\ci{0.6}} & $13.54$\,\ci{1.1} & $36.65$\,\ci{7.4} & \rkc{$11.20$\,\ci{0.6}} \\

\midrule
\multicolumn{6}{c}{\textbf{G1 (68D)}}\\
\midrule
$k_p$    & \rka{$\mathbf{1.11}$\,\ci{0.2}} & \rkc{$3.50$\,\ci{0.2}} & $4.05$\,\ci{0.4} & \rkb{$2.64$\,\ci{0.3}} & $5.16$\,\ci{0.9} \\
$k_d$    & \rka{$\mathbf{2.60}$\,\ci{0.5}} & \rkc{$4.89$\,\ci{0.4}} & $7.88$\,\ci{1.2} & \rkb{$4.17$\,\ci{1.0}} & $23.82$\,\ci{6.0} \\
Base     & \rka{$\mathbf{0.11}$\,\ci{0.0}} & $2.76$\,\ci{0.4} & $3.52$\,\ci{1.5} & \rkb{$0.19$\,\ci{0.1}} & \rkc{$2.54$\,\ci{1.4}} \\
All      & \rka{$\mathbf{1.28}$\,\ci{0.2}} & \rkc{$3.72$\,\ci{0.3}} & $5.15$\,\ci{0.5} & \rkb{$2.33$\,\ci{0.4}} & $10.50$\,\ci{2.6} \\

\midrule
\multicolumn{6}{c}{\textbf{Crazyflie (17D)}}\\
\midrule
$k_f$    & $2.14$\,\ci{0.2} & \rkb{$1.34$\,\ci{0.2}} & \rkc{$1.48$\,\ci{0.5}} & $2.34$\,\ci{0.1} & \rka{$\mathbf{1.30}$\,\ci{0.2}} \\
$k_m$    & \rkb{$3.44$\,\ci{0.6}} & $6.13$\,\ci{1.3} & \rka{$\mathbf{2.64}$\,\ci{1.4}} & $8.15$\,\ci{0.1} & \rkc{$5.73$\,\ci{1.1}} \\
$\tau$   & \rkb{$0.84$\,\ci{0.1}} & \rkc{$0.90$\,\ci{0.1}} & \rka{$\mathbf{0.72}$\,\ci{0.2}} & $1.24$\,\ci{0.1} & $0.91$\,\ci{0.1} \\
$I_d$    & \rkb{$2.04$\,\ci{0.3}} & $2.80$\,\ci{0.4} & \rka{$\mathbf{1.50}$\,\ci{0.5}} & $4.27$\,\ci{0.3} & \rkc{$2.68$\,\ci{0.3}} \\
All      & \rkb{$2.11$\,\ci{0.2}} & $2.79$\,\ci{0.4} & \rka{$\mathbf{1.59}$\,\ci{0.5}} & $4.00$\,\ci{0.1} & \rkc{$2.66$\,\ci{0.4}} \\
\bottomrule
\end{tabular}
\vspace{-8pt}
\end{table}

\begin{figure}[t]
\centering
\includegraphics[width=\columnwidth]{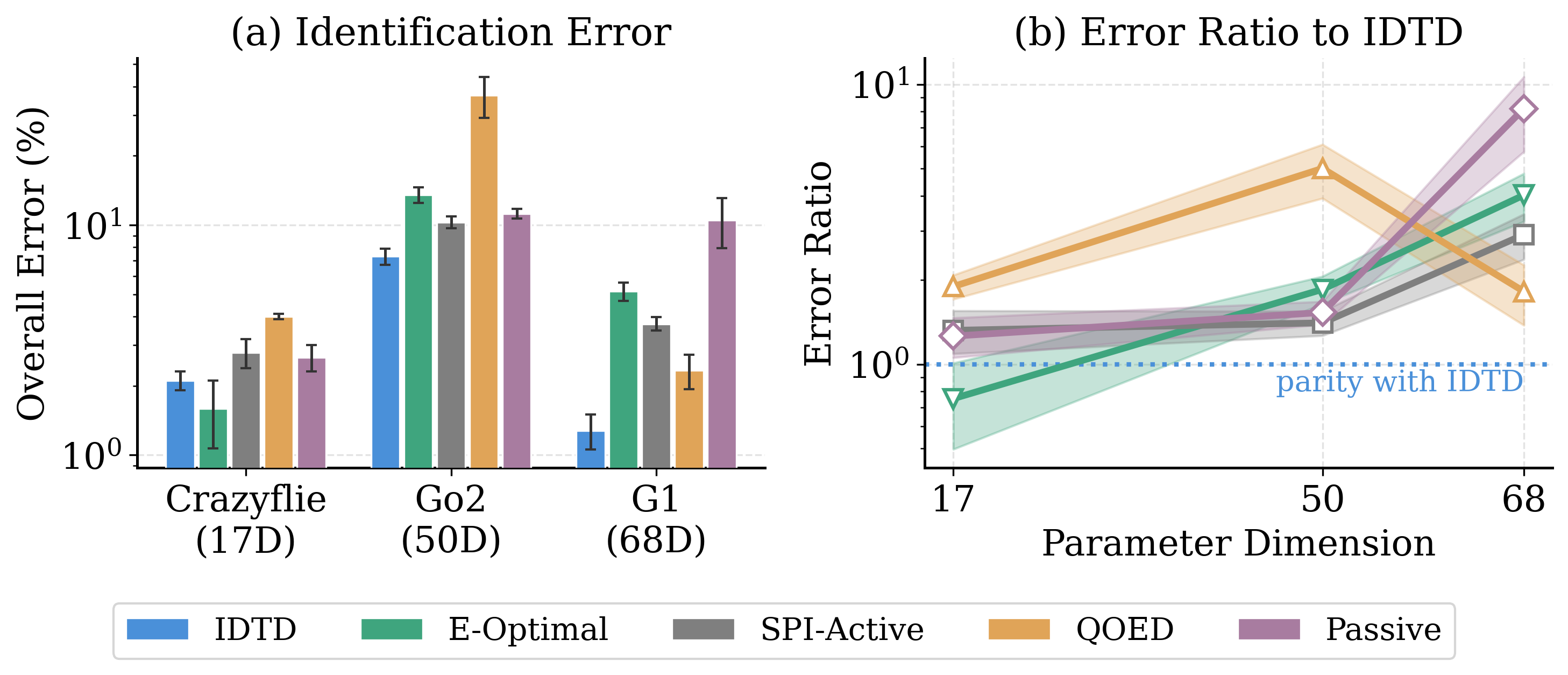}
\vspace{-20pt}
\caption{(a) Overall identification error on each platform. (b) Error ratio of each baseline to \oursol{}, where values above the dotted line indicate higher error than \oursol{}. Shaded bands show $95\%$ confidence intervals.}
\vspace{-12pt}
\label{fig: exp_overall_figure}
\end{figure}

\noindent\textbf{Main results.}
Table~\ref{tab:accuracy_all} compares parameter identification accuracy against the ground-truth parameters on Go2, G1, and Crazyflie.
For each method, we optimize exploration commands for $2$ pretrained policies and run system identification with $5$ seeds each, yielding $10$ runs in total.
We report the absolute parameter error normalized by its search range as a percentage (mean $\pm$ 95\% confidence interval), with shading indicating the top three methods within each parameter group.

Across Go2 and G1, \oursol{} achieves the lowest overall identification error, reducing the error from $10.28\%$ to $7.31\%$ on the Go2 and from $3.72\%$ to $1.28\%$ on the G1 compared with \spiactive{}.
The improvements are particularly pronounced for $k_p$, $k_d$, joint friction, and base inertial parameters, which are easily entangled during locomotion because different combinations of actuator gains, friction, and inertial parameters can produce similar joint motions.
On the Crazyflie, \oursol{} remains competitive at $2.11\%$ overall error, while E-Optimal achieves the lowest error at $1.59\%$.
Figure~\ref{fig: exp_overall_figure} further highlights the stronger relative advantage of \oursol{} on the higher-dimensional identification problems, except against QOED on the G1, whose subset-selection strategy leads to less consistent performance across morphologies.
These results demonstrate that the per-parameter decoupling objective in \oursol{} improves parameter identifiability during the trajectory design stage, enabling reliable estimation across systems with different morphologies and parameter dimensions.

\begin{figure}[t]
\centering
\includegraphics[width=\columnwidth]{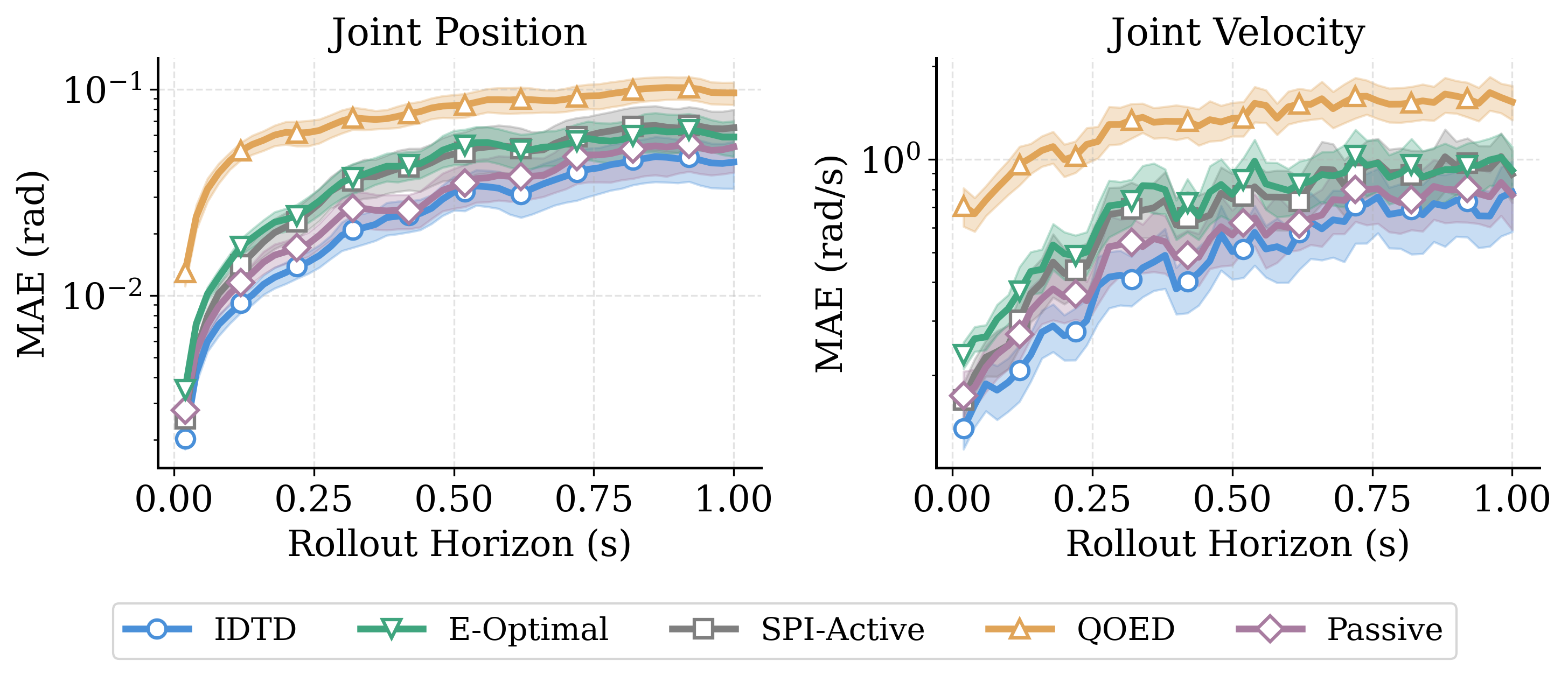}
\vspace{-20pt}
\caption{Forward-dynamics prediction error versus rollout horizon on the Go2. Shaded bands show $95\%$ confidence intervals.}
\vspace{-12pt}
\label{fig:fd_horizon}
\end{figure}

\noindent\textbf{Forward-dynamics fidelity.}
To assess how faithfully the identified parameters reproduce the target system dynamics, we set each method's estimate in simulation and measure forward-dynamics prediction error.
Since a low prediction error on identification trajectories does not necessarily imply accurate identification (Sec.~\ref{subsec:linear_exp}), we evaluate on held-out trajectories that span a broad range of the robot's operating space.
Figure~\ref{fig:fd_horizon} shows the open-loop joint-position and joint-velocity prediction errors as the rollout horizon increases.
\oursol{} achieves the lowest prediction error across rollout horizons, providing further evidence that its estimates describe the target system rather than only the identification trajectories.

\begin{table}[b]
\vspace{-16pt}
\centering
\caption{Downstream transfer performance on the Go2.\\
Mean $\pm$ 95\% confidence interval, darker = higher rank.}
\label{tab:go2_transfer}
\setlength{\tabcolsep}{4.5pt}
\scriptsize
\begin{tabular}{lcccc}
\toprule
Method & Lin.\ (m/s) & Ang.\ (rad/s) & Fall (\%) & Energy (J/m) \\
\midrule
\oursol{}          & \rka{$\mathbf{0.159}$\,\ci{.001}} & \rka{$\mathbf{0.177}$\,\ci{.002}} & \rka{$\mathbf{0.71}$\,\ci{0.26}} & \rka{$\mathbf{242.8}$\,\ci{2.7}} \\
\spiactive{}       & \rkb{$0.186$\,\ci{.002}} & \rkb{$0.187$\,\ci{.002}} & \rkb{$1.81$\,\ci{0.41}} & \rkb{$255.4$\,\ci{2.7}} \\
\textsc{Passive}   & \rkc{$0.190$\,\ci{.002}} & \rkc{$0.234$\,\ci{.003}} & \rkc{$2.59$\,\ci{0.49}} & \rkc{$270.2$\,\ci{3.1}} \\
\textsc{Zero-shot} & $0.364$\,\ci{.004} & $0.433$\,\ci{.006} & $22.35$\,\ci{1.26} & $547.9$\,\ci{7.4} \\
\bottomrule
\end{tabular}
\end{table}

\noindent\textbf{Downstream transfer.}
\label{subsec:transfer}
To evaluate whether improved identification leads to better policy performance under the target dynamics, we conduct a transfer experiment on the Go2. 
Specifically, for each method, we retrain a policy under the identified dynamics and evaluate the resulting policy on the ground-truth target dynamics. 
As the identification errors of E-Optimal and QOED are substantially larger on the Go2 (Table~\ref{tab:accuracy_all}), policy training under their identified dynamics does not yield stable policies, so we exclude them from the transfer evaluation.
We additionally include a zero-shot baseline that directly deploys a policy trained with the nominal simulator parameters.
As shown in Table~\ref{tab:go2_transfer}, \oursol{} achieves the lowest linear and angular velocity tracking errors, fall rate, and energy consumption among all methods.
The lower energy consumption is consistent with improved PD-gain identification reducing unnecessary corrective actuation.
These results show that the parameters identified by \oursol{} translate into better downstream policy transfer to the target dynamics.

\noindent\textbf{Parameter coupling under each objective.}
To analyze parameter coupling, we examine the FIM induced by each exploration method on G1.
%
%
Specifically, for the $(k_p,k_d)$ pair of the wrist-pitch joint, we visualize the $95\%$ confidence ellipse from the corresponding block $C=(F^{-1})_{{ij}}$, which the Cram\'er--Rao bound gives as a lower bound on their covariance.
We additionally report the normalized Fisher correlation $\rho_{ij}=F_{ij}/\sqrt{F_{ii}F_{jj}}$ between $k_p$ and $k_d$ across G1 joints.

\begin{figure}[t]
\centering
\includegraphics[width=\columnwidth]{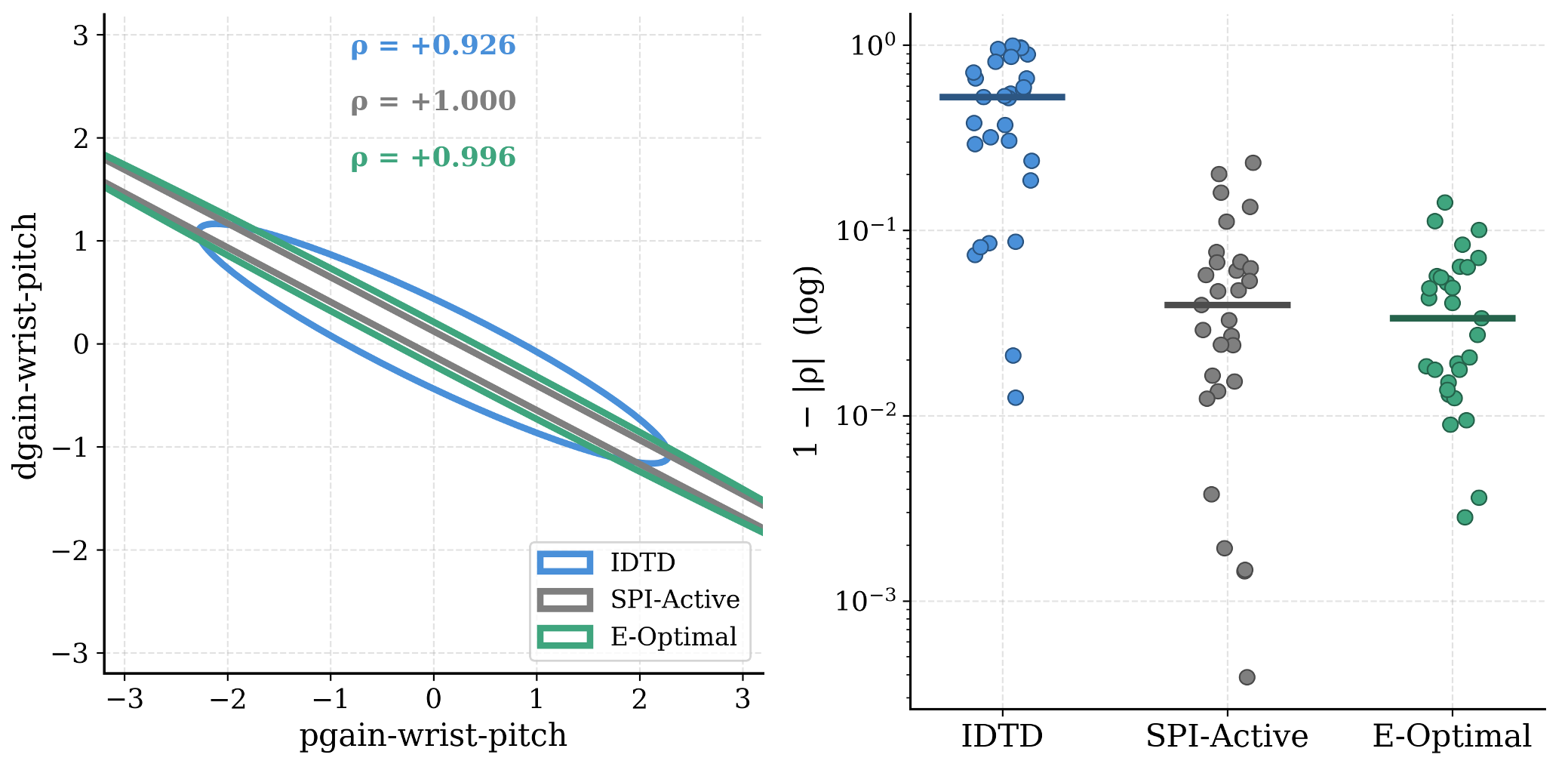}
\vspace{-16pt}
\caption{
Left: covariance ellipse for the $(k_p,k_d)$ pair of the wrist-pitch joint.
Right: $1-|\rho_{ij}|$ for the $(k_p,k_d)$ pair of all G1 joints.}
\vspace{-14pt}
\label{fig: coupling_figure}
\end{figure}

In Figure~\ref{fig: coupling_figure}, \oursol{} produces a substantially more isotropic ellipse, while \spiactive{} and \textsc{E-Optimal} yield highly elongated ellipses.
An elongated ellipse indicates strong parameter coupling because its major axis represents a direction of high uncertainty along which changes in one parameter can be compensated by changes in the other.
Consequently, one parameter can be overestimated while the other is underestimated without being well distinguished by the observed trajectory.
In contrast, the more isotropic ellipse of \oursol{} indicates weaker coupling and therefore constrains the individual parameters to a more localized region.
The same trend holds across the G1 joints, where \oursol{} yields larger $1 - |\rho_{ij}|$ overall for the $(k_p, k_d)$ pairs beyond the representative example.
Although the A-optimal criterion of \spiactive{} also implicitly penalizes this correlation, it does not separate this effect from information quantity, resulting in weaker parameter separation in practice.
This confirms that the decoupling objective of \oursol{} suppresses the parameter correlation responsible for error absorption between parameters, providing a mechanistic explanation for the improved identification accuracy.

\begin{figure}[b]
\centering
\includegraphics[width=\columnwidth]{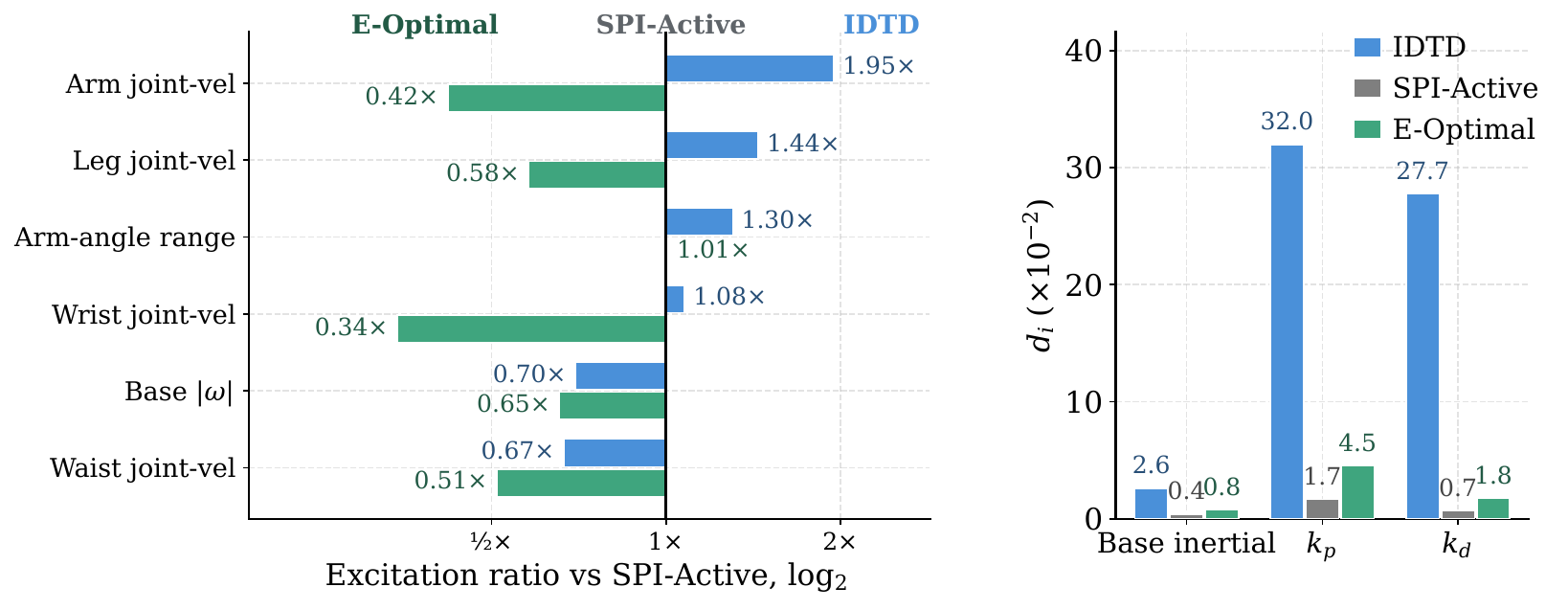}
\vspace{-16pt}
\caption{Left: the excitation of each kinematic channel relative to \spiactive{} on a $\log_2$ scale. Right: the per-group mean of the decoupling factor $d_i$.}
\vspace{-15pt}
\label{fig: behavior_analysis}
\end{figure}

\noindent\textbf{Excitation allocation.}
To characterize how different objectives allocate excitation, we train the full G1 policy, rather than optimizing its commands, using each method's objective as the policy reward.
In Figure~\ref{fig: behavior_analysis}, the left panel reports the excitation of each kinematic channel relative to \spiactive{}, and the right panel reports the decoupling factor $d_i$ from Eq.~\eqref{eqn: decoupling_factor}.

Compared with \spiactive{}, \oursol{} produces substantially stronger limb motion, increasing arm and leg joint velocities to $1.95\times$ and $1.44\times$, respectively, while expanding the arm-angle range to $1.30\times$.
\textsc{E-Optimal} instead falls below \spiactive{} on every channel, which suggests that its minimum-eigenvalue objective provides little incentive to increase excitation in already informative directions.
\spiactive{}, on the other hand, induces the strongest base and waist motion among the three methods, consistent with its objective, which can favor aggressive base motion as an easy way to lower the aggregate estimation uncertainty even when the parameter effects remain coupled.
The right panel shows that \oursol{} achieves substantially higher decoupling factors across all three parameter groups.
\textsc{E-Optimal} also improves $d_i$ over \spiactive{}, but, as shown in the left panel, does so under weaker excitation.
Together, the two panels show that \oursol{} combines targeted excitation with strong parameter decoupling, whereas \spiactive{} favors strong excitation with weak decoupling and \textsc{E-Optimal} improves decoupling but with limited excitation.

\begin{table}[t]
\vspace{-16pt}
\centering
\caption{Ablation of \oursol{} design choices on the Go2 (50D). \\ 
Mean $\pm$ 95\% confidence interval, darker = higher rank.}
\label{tab:ablation}
\scriptsize
\begin{tabular}{lcccc}
\toprule
Group & \textbf{\oursol{}} & \textbf{Mean agg.} & \textbf{Unidelta} & $\boldsymbol{\alpha=0}$ \\
\midrule
$k_p$    & \rka{$\mathbf{5.23}$\,\ci{0.7}} & \rkc{$8.18$\,\ci{1.0}} & \rkb{$7.39$\,\ci{0.4}} & $16.30$\,\ci{0.9} \\
$k_d$    & \rka{$\mathbf{6.71}$\,\ci{0.7}} & \rkc{$11.72$\,\ci{1.5}} & \rkb{$9.05$\,\ci{0.7}} & $14.85$\,\ci{1.8} \\
Joint    & \rka{$\mathbf{10.99}$\,\ci{1.4}} & \rkc{$19.63$\,\ci{1.5}} & \rkb{$11.03$\,\ci{0.7}} & $51.41$\,\ci{2.7} \\
Foot     & \rkb{$5.90$\,\ci{0.9}} & \rkc{$7.31$\,\ci{1.0}} & \rka{$\mathbf{4.29}$\,\ci{0.5}} & $15.09$\,\ci{0.8} \\
Base     & \rkb{$7.71$\,\ci{1.3}} & \rka{$\mathbf{5.51}$\,\ci{2.3}} & \rkc{$9.81$\,\ci{1.2}} & $15.77$\,\ci{1.9} \\
All      & \rka{$\mathbf{7.31}$\,\ci{0.6}} & \rkc{$10.47$\,\ci{0.9}} & \rkb{$8.31$\,\ci{0.4}} & $22.68$\,\ci{0.9} \\
\bottomrule
\end{tabular}
\end{table}

\noindent\textbf{Ablation.}
In Table~\ref{tab:ablation}, we ablate three design choices on the Go2.
\textbf{\textit{Mean agg.}} replaces the $\max$ over segments in Eq.~\eqref{eqn: coverage_credit} and the logarithmic sum over parameters in Eq.~\eqref{eqn: coverage_objective} with plain averages, removing mechanisms that promote coverage of weakly covered parameters.
\textbf{\textit{Unidelta}} removes the sensitivity normalization of Sec.~\ref{subsec: normalization} and perturbs every parameter by a common absolute step.
$\boldsymbol{\alpha=0}$ removes the decoupling normalization, so the score reverts to the raw Schur complement.

\textbf{\textit{Mean agg.}} increases the overall error from $7.31\%$ to $10.47\%$ and particularly degrades joint-friction identification, as weaker coverage pressure concentrates excitation on the base.
\textbf{\textit{Unidelta}} overweights the foot-friction parameters under a common perturbation scale, increasing the overall error from $7.31\%$ to $8.31\%$.
$\boldsymbol{\alpha=0}$ is most damaging, raising the error to $22.68\%$, as the objective becomes dominated by absolute information magnitude rather than parameter decoupling.
These results clarify the role of each component under uneven parameter coverage, sensitivity, and information magnitude.



\subsection{Real-world Experiments}
\label{subsec: sim2real}

\noindent\textbf{Setup.}
We deploy \oursol{} on a real Booster K1 humanoid to validate its effectiveness for real-world system identification.
Our hardware setup uses onboard proprioceptive sensing from joint encoders and an IMU, without external motion capture.
We therefore focus on actuator- and joint-level parameters rather than floating-base inertial properties such as the CoM and body inertia.
Specifically, we identify a 60-dimensional parameter space consisting of actuator torque-lag time constants (8D), which capture first-order torque-response delay; velocity-dependent actuator gains (8D), which model transmission losses during joint motion; joint-friction coefficients (22D); and joint armatures (22D), which represent reflected rotor inertia.
For the actuator parameters, we group corresponding left- and right-side joints by motor class and limb segment, resulting in eight groups for both torque lag and dynamic gain.
We use the Newton physics simulator~\cite{newton} for forward simulation and parameter identification.
The state components used in the identification objective in Eq.~\eqref{eqn: sysid_objective} are the base orientation and angular velocity from the onboard IMU and the joint positions and velocities from the encoders.

\begin{table}[b]
\centering
\caption{Open-loop prediction on real-world K1 trajectories. \\
Mean $\pm$ 95\% confidence interval; darker = higher rank.}
\label{tab:k1}
\setlength{\tabcolsep}{1pt}
\scriptsize
\begin{tabular}{lccccc}
\toprule
Method & Cost & $q$ (deg) & $\dot q$ (rad/s) & $\omega$ (rad/s) & Orient. (deg) \\
\midrule
Nominal & $2.291$ & $1.146$ & $0.519$ & $0.322$ & $1.247$ \\
\textsc{Passive} & \rkb{$2.018${\tiny$\pm.003$}} & \rkb{$1.095${\tiny$\pm.001$}} & \rkb{$0.482${\tiny$\pm.001$}} & \rkb{$0.314${\tiny$\pm.001$}} & \rkb{$1.242${\tiny$\pm.004$}} \\

\spiactive{} & \rkc{$2.061${\tiny$\pm.005$}} & \rkc{$1.108${\tiny$\pm.004$}} & \rkc{$0.489${\tiny$\pm.001$}} & \rka{$\mathbf{0.312}${\tiny$\pm.001$}} & \rkc{${1.252}${\tiny$\pm.003$}} \\

\oursol{} & \rka{$\mathbf{1.961}${\tiny$\pm.005$}} & \rka{$\mathbf{1.074}${\tiny$\pm.002$}} & \rka{$\mathbf{0.473}${\tiny$\pm.001$}} & \rkb{$0.314${\tiny$\pm.001$}} & \rka{$\mathbf{1.233}${\tiny$\pm.005$}} \\
\bottomrule
\end{tabular}
\vspace{-12pt}
\end{table}

\noindent\textbf{Results.} 
We evaluate the identified parameters by measuring open-loop prediction error on held-out trajectories that were not used for identification.
Table~\ref{tab:k1} reports the weighted prediction cost from Eq.~\eqref{eqn: sysid_objective}, together with the prediction errors for joint position $q$, joint velocity $\dot q$, base angular velocity $\omega$, and base orientation.
All three methods improve over the nominal model, showing that system identification recovers parameters that better capture real-world dynamics.
\oursol{} attains the largest cost reduction at $14.4\%$, compared with $11.9\%$ for \textsc{Passive} and $10.0\%$ for \spiactive{}, and achieves the lowest error on most state components.
Together, these results demonstrate the applicability of \oursol{} to real-world system identification on humanoid hardware.
Real-world exploration videos are provided in the supplementary video.

\section{Conclusion}
We presented \oursol{}, a trajectory design framework for reliable sampling-based system identification that explicitly accounts for parameter separability during data collection.
\oursol{} combines a normalized Schur-complement score with segment-wise coverage and logarithmic aggregation to balance parameter separability and information quantity.
Across simulated Go2, G1, and Crazyflie platforms, \oursol{} improved identification accuracy, dynamics prediction, and parameter decoupling, while on the Booster K1 humanoid the identified parameters predicted unseen real trajectories more accurately.

\noindent\textbf{Limitations and future work.}
While \oursol{} improves forward-dynamics prediction on the real K1, downstream control improvements remain limited, possibly due to the restricted parameter space identifiable from onboard sensing.
Future work could use more accurate state estimation, such as external motion capture, to identify a broader parameter set and improve downstream transfer.

\bibliographystyle{IEEEtran}
\bibliography{references}

\appendix

\section{Implementation Details}
\label{app:implementation}

\subsection{Trajectory Optimization}
\label{app:trajectory_optimization}

For all platforms, we optimize the exploration command sequence using
CMA-ES. Each trajectory is divided into $M=5$ segments of $2$ seconds,
and the command within each segment is represented by a degree-$5$
B\'ezier curve with six control points per command channel.
CMA-ES is initialized with $\sigma_0=0.25$ and run for at most $3000$
generations, with early stopping when the objective variation falls
below $10^{-3}$.
The initial mean is obtained by fitting the B\'ezier control points to
the center of the admissible command range.
The population sizes are $20$, $12$, and $20$ for the Go2, G1, and
Crazyflie, respectively.
The corresponding command dimensions are $9$, $3$, and $2$.
Table~\ref{tab:active_optimization_details} summarizes the optimization
settings.

\begin{table}[t]
\centering
\caption{Trajectory-optimization settings used in the simulation experiments.}
\label{tab:active_optimization_details}
\setlength{\tabcolsep}{5pt}
\begin{tabular}{lccc}
\toprule
Setting & Go2 & G1 & Crazyflie \\
\midrule
SysID dimension $P$       & 50   & 68   & 17 \\
Command dimension $C$     & 9    & 3    & 2 \\
CMA-ES population         & 20   & 12   & 20 \\
Maximum generations       & 3000 & 3000 & 3000 \\
Initial $\sigma_0$        & 0.25 & 0.25 & 0.25 \\
Segments $M$              & 5    & 5    & 5 \\
B\'ezier degree           & 5    & 5    & 5 \\
Segment duration (s)      & 2.0  & 2.0  & 2.0 \\
Control timestep (s)      & 0.02 & 0.02 & 0.01 \\
Parallel groups           & 8    & 8    & 8 \\
Synchronization interval  & 5    & 5    & 5 \\
Schur regularization      & $10^{-6}$ & $10^{-6}$ & $10^{-6}$ \\
Coverage $\varepsilon$    & $10^{-6}$ & $10^{-6}$ & $10^{-6}$ \\
\bottomrule
\end{tabular}
\end{table}

\subsection{Fisher-Information Estimation}
\label{app:fim_estimation}

We estimate parameter sensitivities by finite differences using one
nominal rollout and one perturbed rollout for each SysID parameter.
For each trajectory candidate, we evaluate eight parallel groups, each
containing the nominal environment and the $P$ parameter-perturbed
environments.
All eight groups share the same parameter vector used for Fisher
information evaluation, providing repeated estimates of the
finite-difference sensitivities.
To limit divergence between the nominal and perturbed rollouts, the
perturbed environments are synchronized with the nominal environment
every five simulation steps.
The resulting sensitivities are accumulated within each trajectory
segment to construct the segment-wise Fisher information matrices
$F^{(s)}$ used by \oursol{}.

For numerical stability, when computing the Schur complement in
Eq.~\eqref{eqn: schur_single}, we regularize the nuisance-parameter
block as
\begin{equation}
S_i(F)
=
F_{ii}
-
F_{i,\setminus i}
\left(
F_{\setminus i,\setminus i}
+
\lambda_{\mathrm{Schur}} I
\right)^{-1}
F_{\setminus i,i},
\end{equation}
where $\lambda_{\mathrm{Schur}}=10^{-6}$.
We additionally use $\varepsilon=10^{-6}$ in the logarithmic coverage
objective of Eq.~\eqref{eqn: coverage_objective}.

\subsection{Decoupling Exponents}
\label{app:decoupling_exponents}

The decoupling exponent $\alpha_i$ in
Eq.~\eqref{eqn: decoupling_factor} is assigned by parameter group and
held fixed throughout each experiment.
As discussed in Sec.~\ref{subsec: normalization}, larger values of
$\alpha_i$ more strongly remove the contribution of absolute
information magnitude and place greater emphasis on the pure
decoupling ratio, whereas smaller values retain more dependence on
excitation strength.
Across all experiments, we use $\alpha_i\in[0.7,1.0]$.
The values used for each parameter group are reported in
Table~\ref{tab:alpha_values}.

\begin{table}[t]
\centering
\caption{Decoupling exponents $\alpha_i$ used for each parameter group.}
\label{tab:alpha_values}
\setlength{\tabcolsep}{6pt}
\begin{tabular}{llc}
\toprule
Platform & Parameter group & $\alpha_i$ \\
\midrule
\multirow{4}{*}{Go2}
 & Base mass/inertial & 1.00 \\
 & Actuator PD        & 0.90 \\
 & Foot friction      & 0.70 \\
 & Contact            & 0.75 \\
\midrule
\multirow{3}{*}{G1}
 & Base inertial      & 0.80 \\
 & $k_p$              & 1.00 \\
 & $k_d$              & 0.75 \\
\midrule
\multirow{3}{*}{Crazyflie}
 & Motor              & 1.00 \\
 & Inertial           & 0.85 \\
 & CoM $x,y$          & 0.70 \\
\bottomrule
\end{tabular}
\end{table}

\subsection{Finite-Difference Perturbations}
\label{app:finite_difference}

We use parameter-group-specific finite-difference perturbations
$\delta\theta_i$ according to the numerical scale of each physical
parameter.
The perturbation values are fixed within each parameter group and are
used consistently throughout trajectory optimization.
Table~\ref{tab:sim_fd} reports the parameter ranges and finite-difference
steps used for all three simulated platforms.

\begin{table}[t]
\centering
\caption{Parameter ranges and finite-difference perturbations used in simulation.}
\label{tab:sim_fd}
\setlength{\tabcolsep}{3.0pt}
\small
\begin{tabular}{llccc}
\toprule
Platform & Parameter & Dim. & Range & $\delta\theta_i$ \\
\midrule
\multirow{5}{*}{Go2}
& Base inertial
    & 10 & Nominal $\pm$ margin & 0.10 \\
& $k_p$ (N\,m/rad)
    & 12 & $[5.0,100.0]$ & 2.00 \\
& $k_d$ (N\,m\,s/rad)
    & 12 & $[0.1,10.0]$ & 0.10 \\
& Joint friction (N\,m)
    & 12 & $[0.0,5.0]$ & 0.05 \\
& Foot friction
    & 4 & $[0.1,2.0]$ & 0.05 \\
\midrule
\multirow{3}{*}{G1}
& Base inertial
    & 10 & Nominal $\pm$ margin & 0.10 \\
& $k_p$ (N\,m/rad)
    & 29 & $[8.0,110.0]$ & 2.00 \\
& $k_d$ (N\,m\,s/rad)
    & 29 & $[0.3,10.0]$ & 0.10 \\
\midrule
\multirow{5}{*}{Crazyflie}
& $k_f$ ($10^{-10}$ N/rpm$^2$)
    & 4 & $[2.2,4.1]$ & 0.08 \\
& $k_m$ ($10^{-12}$ N\,m/rpm$^2$)
    & 4 & $[4.0,16.0]$ & 0.25 \\
& $\tau$ ($10^{-2}$ s)
    & 4 & $[2.0,15.0]$ & 0.40 \\
& $I_{\mathrm{diag}}$ ($10^{-5}$ kg\,m$^2$)
    & 3 & $[0.5,4.5]$ & 0.05 \\
& CoM $(x,y)$
    & 2 & $[-8.0,8.0]$ & 0.40 \\
\bottomrule
\end{tabular}
\end{table}

\subsection{Real-World K1 Experiment}
\label{app:k1_experiment}

For the real-world Booster K1 experiments, we identify a
$60$-dimensional parameter vector consisting of $8$ torque-lag time
constants, $8$ velocity-dependent actuator gains, $22$ joint-friction
coefficients, and $22$ joint-armature parameters.
Table~\ref{tab:k1_parameter_ranges} summarizes the parameter ranges,
finite-difference perturbations, and decoupling exponents used for
trajectory design and system identification.
Parameter fitting is performed using CMA-ES with a population size of
$100$, an initial standard deviation of $0.5$, and a function-tolerance
stopping criterion of $10^{-6}$.
The identification objective additionally includes a MAP-style
regularization term that penalizes deviations from the nominal
parameters, with parameter-group-specific regularization strengths
ranging from $10^{-3}$ to $5\times10^{-2}$.

\begin{table}[t]
\centering
\caption{Parameterization and optimization settings for the real Booster K1.}
\label{tab:k1_parameter_ranges}
\setlength{\tabcolsep}{3.5pt}
\small
\begin{tabular}{lcccc}
\toprule
Parameter & Dim. & Range & $\delta\theta_i$ & $\alpha_i$ \\
\midrule
Torque lag ($10$ ms)
    & 8 & $[0,15]$ & 0.50 & 0.80 \\
Dynamic gain
    & 8 & $[0.1,1.0]$ & 0.02 & 0.70 \\
Joint friction
    & 22 & $[0,0.8]$ & 0.02 & 0.70 \\
Armature ($10^{-2}$)
    & 22 & $[0.05,15]$ & 0.10 & 1.00 \\
\bottomrule
\end{tabular}
\end{table}

\end{document}